\pdfoutput=1

\documentclass[11pt]{article}

\usepackage[]{EMNLP2023}

\usepackage{times}
\usepackage{latexsym}

\usepackage{amsmath}
\usepackage{url}

\usepackage{booktabs}
\usepackage{graphicx}
\usepackage{amsmath}
\usepackage{amssymb}
\usepackage{multirow}
\usepackage{verbatim}
\usepackage{color}
\usepackage{empheq}
\usepackage{comment}
\usepackage{ulem}

\usepackage[T1]{fontenc}

\usepackage[utf8]{inputenc}

\usepackage{microtype}

\usepackage{inconsolata}

\title{GOVERN: Gradient Orientation Vote Ensemble for Multi-Teacher
Reinforced Distillation}

\author{
Wenjie Zhou$^{1}$\thanks{~Corresponding author.}\space\space\space
Zhenxin Ding$^{1}$\space\space\space
Xiaodong Zhang$^{1}$\space\space\space
\\
{\bf
Haibo Shi$^{1}$\space\space\space
Junfeng Wang$^{1}$\space\space\space
Dawei Yin$^{1}$\space\space\space
}\\  
$^1$Baidu Inc., Beijing, China\\
\texttt{wjzhou013@pku.edu.cn}~~zhenxinding@gmail.com~~zxdcs@pku.edu.cn \\
\texttt{haiboshi@outlook.com~~wangjunfeng@baidu.com~~yindawei@acm.org} \\
}

\begin{document}

\maketitle

\begin{abstract}
Pre-trained language models have become an integral component of question-answering systems, achieving remarkable performance. However, for practical deployment, it is crucial to perform knowledge distillation to maintain high performance while operating under computational constraints. In this paper, we address a key question: given the importance of unsupervised distillation for student model performance,  how can knowledge from multiple teacher models be effectively ensemble during this stage without the guidance of labels? We propose a novel algorithm, GOVERN, to tackle this issue. GOVERN has demonstrated significant improvements in both offline and online experiments, enabling the student model to achieve results comparable to that of teacher ensembles. Our experiments show that GOVERN remarkably requires a mere 1\% of the ensemble method's inference budget to achieve 99.5\% of performance. The proposed algorithm has been successfully deployed in a real-world commercial question-answering system, demonstrating its real-world applicability.

\end{abstract}

\section{Introduction}

Traditional search engine aims at deliver relevant web pages to satisfy users' question, while sometimes the single paragraph that answer the question might buried deep in a web page, it asks for a web-based Open domain Question Answering (OpenQA) system to find that needle-in-a-haystack info (e.g. \citealp{qu-etal-2021-rocketqa, zhang-etal-2023-survey-efficient}).


BERT-liked pre-trained language models have achieved state-of-the-art performance in OpenQA (e.g. \citealp{zhang2021poolingformer}). However, due to computational costs, the direct application of these models in real-time search engines like Google is currently unfeasible. For instance, the top-performing models on the Natural Question dataset, R2-D2 \citep{fajcik-etal-2021-r2-d2} and UnitedQA \citep{cheng-etal-2021-unitedqa} come with 1.29B and 2.09B model parameters. Further complicating matters is the fact that ensemble methods, which can enhance performance, entail even greater computational overheads.

The distillation of knowledge from multiple teachers has emerged as a powerful technique for improving the performance and generalization of DNN while reducing the computational cost. This two-stage training paradigm, which training large model with limited labeled data as teacher and then using it to generate soft label on large amount unlabeled data for the purpose of student training, was first proposed by \citet{hinton2015distilling}. Since the knowledge from single teacher may be biased and inaccurate, ensemble distillation from multiple teachers was considered by previous works to achieve more promising performance (e.g. \citealp{You2017LearningFM, Fukuda2017EfficientKD}).

Several dynamic distillation methods were proposed to solve the problem that different teacher is good at different sample and low-quality teachers may mislead the student. e.g. \citet{DBLP:conf/aaai/YuanSPLGFJ21} proposed a novel RL-based approach to dynamically assigns weights among teachers, \citet{cai-etal-2022-pile} ensembles multi-teacher logits supervised by human-annotated labels in an iterative way. But these dynamic teacher selection methods need supervision signal as guidance, that means they can not apply to unsupervised distillation which is the most important stage in distillation \citep{DBLP:journals/corr/abs-2106-02241}.

In this paper, we propose \textbf{Gradient Orientation Vote Ensemble Reinforced distillatioN} (\textbf{GOVERN}) to do sample-wise dynamic teacher selection without the need of label guidance. 

Our main contributions are summarized as follows:

\begin{itemize}
    \item We propose GOVERN to do sample-wise dynamic teacher selection without the need of label guidance. We also give a proof that GOVERN can perform better than mean ensemble. To the best of our knowledge, GOVERN is the first method which can find sample-wise high-quality teachers without label guidance.
    \item We propose a novel distillation framework for industrial applications that integrates the GOVERN method into both unsupervised and supervised distillation stages. This framework enhances the performance of student neural networks, enabling them to achieve results comparable to those of ensemble methods. The potential benefits of this approach make it a valuable contribution to industrial OpenQA systems.
    \item Extensive experiments show that GOVERN is benefit in both distillation stage and can boost the real-world question answering system.
\end{itemize}

\section{Answer Selection Task}
In a web-based Open domain Question Answering (OpenQA) system, the primary objective is to select the relevant paragraphs $A_q = {a_i}_{i=1}^{N} \subset P_q$ which can solve the custom's question $q \in Q$, where $P_q$ is a collection of paragraphs obtained in web pages retrieved by search engine. A classic framework of this system is made up of two-stage modules including retriever and ranker, where both modules can be distilled down to a task of classifying the relevance between a question and an answer. Our work focus on improving the performance of classification model with the limit of model size.

The classification model assesses the relevance of a paragraph, denoted as $p$, to a specific question, denoted as $q$, by calculating the relevance score, $f(q, p; \theta)$. This scoring function, $f$, which is parameterized by $\theta$, symbolizes the degree of relevance between the question $q$ and the paragraph $p$. In practical application, a score threshold is established for the purpose of classification.

During training, the classification model is optimized by minimizing the loss over training data:
\begin{align}
\label{Loss_classificatin}
\mathop{min} \limits_{\theta} \mathop{\Sigma} \limits_{q \in Q} \mathop{\Sigma} \limits_{p \in P_q}l(y_p^q, f(q, p; \theta))
\end{align}

where $l$ is the loss function such as cross-entropy loss, margin loss or MSE loss, and $y_p^q$ is the relevance label of q-p pair.

\section{Methodology}
We use multiple teachers ensemble distillation as the method to improving the performance of online model with the constriction of computational cost. Within a frequently employed Knowledge Distillation (KD) framework, a large teacher model, denoted as T, is meticulously pretrained or finetuned well ahead of time. The knowledge contained within the teacher model is subsequently transferred to a smaller student model, denoted as S, by minimizing the disparity between the two. This process can be mathematically formulated:
\begin{align}
\label{distillation_loss}
\mathop{min} \limits_{\theta} \mathop{\Sigma} \limits_{x} l(f^S(x; \theta), f^T(x; \Theta))
\end{align}

where x embodies the input sample, while $f^S(\cdot)$ and $f^T(\cdot)$ denote the scoring function of the teacher and student models respectively. Additionally, $L(\cdot)$ serves as a loss function that calculates the variation between the behaviors of the two models. 

\begin{figure*}
\centering
\includegraphics[scale=0.35]{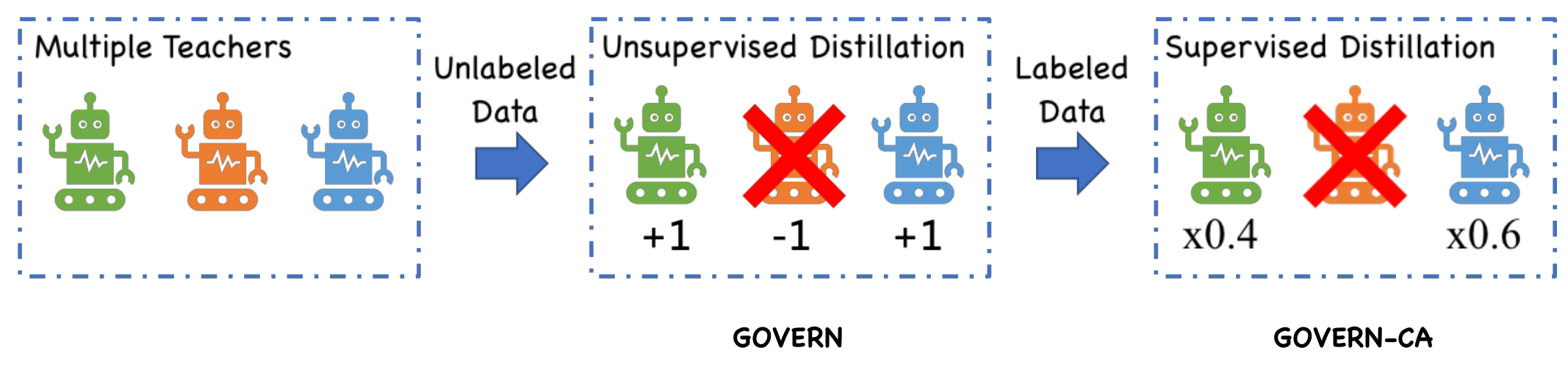}
\caption{\label{unified model} Procedures of Gradient Orientation Vote Ensemble Reinforced Distillation}
\end{figure*}

Specifically, we first utilize unsupervised distillation on a vast amount of task-specific, unlabeled data, followed by supervised distillation on the labeled data. The procedures of the distillation can be viewed in Figure \ref{unified model}.

\subsection{Unsupervised Distillation}
Unsupervised distillation, performed on a substantial amount of task-specific and unlabeled data, is vital for enhancing the performance of the student network. However, due to the absence of supervised signals, the prevalent unsupervised ensemble distillation method resorts to mean-ensemble to amalgamate the abilities of multiple models (\citealp{You2017LearningFM}). 
Other studies have employed a weighted approach whereby individual teacher models are assigned varying weights to accentuate the contribution of higher performing models to knowledge transfer (e.g. \citealp{DBLP:conf/interspeech/FukudaSKTCR17,DBLP:conf/icassp/KwonNLK20,Du2020AgreeTD,DBLP:journals/ijon/LiuZW20}). Methods to determine these weighting coefficients encompass weighting based on experience, calculating the weights based on logistic regression model, latent factor or multi-objective optimization in the gradient space.

While these weighting methods do account for the performance differences among various teachers, they employ a uniform weighting coefficient for all samples during the distillation process. This approach neglects the varying emphasis on each teacher’s abilities and their respective confidence levels regarding 
different samples. 

Here, we propose a novel unsupervised voting method called \textbf{Gradient Orientation Vote Ensemble Reinforced distillatioN} (\textbf{GOVERN}), which does not rely on any human-annotated signals and dynamically assigns different teachers to different samples. In the following, we will introduce the implementation of this unsupervised distillation method and then mathematically prove its superiority over the mean-ensemble method.

It is noted that previous works like UniKD(\citet{DBLP:journals/corr/abs-2204-00548}) and wVID(\citet{DBLP:conf/nips/IliopoulosKBMTV22}) have explored the dynamic assignment of weights. But these methods are used to to evaluate the significance of unlabeled examples, rather than assessing the importance of teachers. These methods could be synergistically integrated with the GOVERN framework, as they enhance unsupervised distillation from distinct perspectives.

\subsubsection{GOVERN}
In unsupervised distillation using mean-ensemble, for a sample, the distilled-model calculates $logit_0$, and N teacher-models calculate $logit_i$ respectively $(1 <= i <= N)$. The distillation loss is:
\begin{align}
Dist(logit_0, Mean(logit_1, ..., logit_N))
\end{align}

where $Dist$ is a distance metric function that can be selected from MSE, cross-entropy, etc.

We take each teacher's gradient descent orientation into consideration while doing ensemble. Specifically, when $logit_i > logit_0$, the gradient of $Dist(logit_0, logit_i)$ calculated is greater than 0, otherwise it is less than 0, so the gradient descent orientation is noted as:
\begin{align}
\label{Grad_i}
Grad_i & = SIGN(gradient(logit_0, logit_i)) \notag \\
 & = \left\{
\begin{array}{rcl}
1       &      & {logit_i      >      logit_0}\\
0       &      & {logit_i      =      logit_0} \\
-1       &      & {logit_i      <      logit_0}
\end{array} \right.
\end{align}

The voted result is calculated as:
\begin{align}
\label{Grad}
\chi(sample)=\left\{
\begin{array}{rcl}
1       &      & {\Sigma_{i=1}^{N}{Grad_i} > 0}\\
0       &      & {\Sigma_{i=1}^{N}{Grad_i} = 0} \\
-1      &      & {\Sigma_{i=1}^{N}{Grad_i} < 0}
\end{array} \right.
\end{align}

Each teacher is considerate as a voter in this way, then the loss for unsupervised distillation is represented as below:
\begin{align}
\label{unsupervised loss}
W_i & = \left\{
\begin{array}{rcl}
1       &      & {\chi * Grad_i \ge 0}\\
0       &      & {\chi * Grad_i < 0} \\
\end{array} \right. \\
\mathcal{L}_{UD} & = MSE(logit_0, \frac{\sum_{i=1}^{N}W_i logit_i}{\sum_{i=1}^{N}{W_i}})
\end{align}
that means, we restrict our approach to guiding the student model’s training under the current sample solely by utilizing the majority of teacher models with consistent gradient orientations.

In Appendix \ref{sec:appendix}, we give a prove that the sample-wise dynamic weighting ensemble algorithm GOVERN is better than mean-ensemble.

\subsection{Supervised Distillation: GOVERN-CA}
Inspired by Conﬁdence-Aware Multi-teacher Knowledge Distillation (CA-MKD) proposed by \citet{DBLP:conf/icassp/ZhangCW22}, we further develop GOVERN algorithm with the help of human label. On each training sample, we select the teachers which share the same gradient descent orientation with the human label. Furthermore, we assign weights among these selected teachers to reflect their sample-wise confidence by calculating the cross entropy loss between the prediction of teachers and human label:
\begin{align}
\label{supervised loss}
y(sample) = \left\{
\begin{array}{rcl}
1,      &      & {if \quad positive}\\
-1,       &      & {if \quad negative} \\
\end{array} \right. \\
W_i = \left\{
\begin{array}{rcl}
1       &      & {y * Grad_i > 0}\\
0       &      & {y * Grad_i \le 0} \\
\end{array} \right. \\
\omega_i = \frac{W_i}{\Sigma_{j}W_j}(1 - \frac{exp(L_{CE}^{i})}{\Sigma_{j}{W_j}exp(L_{CE}^{j})})
\end{align}
where $L^i_{CE}$ denotes the cross entropy loss between the prediction of $i$-th teacher and human label, $Grad_i$ is defined in (\ref{Grad}). The loss for supervised distillation is aggregated with calculated weights:
\begin{align}
\mathcal{L}_{SD} & = MSE(logit_0, \Sigma_{i=1}\omega_i logit_i)
\end{align}

Thereby, we only select teacher with the correct gradient descent orientation. Besides, the teacher whose prediction closely align with the ground-truth labels is assigned a greater weight $\omega_i$. This weighting is attributed to the model’s substantial confidence in making accurate judgments, thereby providing correct guidance.

\begin{table}[ht]
    \centering
    \scalebox{0.75}{
    \begin{tabular}{l|c|c}
    \hline
    \hline
    Dataset & \#Question & \#Question-Paragraph Pair \\
    \hline
        unlabeled data & 3,126,132 & 100M \\
        train data & 190,211 & 2,472,749 \\
        test data & 3,301 & 93,446 \\
    \hline
    \hline
    \end{tabular}
    }
    \caption{Dataset Statistic}
    \label{dataset}
\end{table}

\begin{table*}[!ht]
\centering
\resizebox{\textwidth}{!}{
\begin{tabular}{l@{\quad}|c@{\quad}c@{\quad}c@{\quad}c@{\quad}|c@{\quad}c@{\quad}}
\toprule
\textbf{Model} & \multicolumn{4}{c}{\textbf{Architecture}} & \multicolumn{2}{c}{\textbf{Results}} \\
\textbf{} & \textbf{$n_{params}$} & \textbf{$n_{layers}$} & \textbf{$d_{model}$} & {$n_{heads}$} & {q R@P=90\%} & {qp R@P=90\%} \\
\midrule
\midrule
Teacher1-125M & 125M & 12 & 768 & 12 & 79.51\% & 70.52\% \\
Teacher2-350M & 350M & 24 & 1024 & 16 & 81.79\% & 73.92\% \\
Teacher3-1.5B & 1.5B & 48 & 1536 & 24 & 82.55\% & 73.09\% \\
Teacher4-10B & 10B & 48 & 4096 & 64 & 83.06\% & 73.31\% \\
\midrule
\midrule
\multicolumn{7}{l}{\textbf{Ensemble Model}} \\
\midrule
Mean Ensemble & - & - & - & - & 84.16\% & 76.71\% \\
Logistic Regression Weighted Ensemble & - & - & - & -  & 83.44\% & 76.91\% \\
\midrule
\midrule
\multicolumn{7}{l}{\textbf{Distilled Model}} \\
\midrule
Mean Ensemble Distillation on unlabeled data & 125M & 12 & 768 & 12 & 82.04\% (0.07) & 74.63\% (0.12) \\
LR Ensemble Distillation on unlabeled data & 125M & 12 & 768 & 12 & 81.98\% (0.11) & 75.24\% (0.12) \\
\midrule
GOVERN on unlabeled data & 125M & 12 & 768 & 12 & \textbf{83.65\%} (0.08) & \textbf{76.02\%} (0.14) \\
 + CA-MKD on labeled data & 125M & 12 & 768 & 12 & 82.68\% (0.03) & 75.67\% (0.05) \\
 + GOVERN-CA on labeled data & 125M & 12 & 768 & 12 & \uline{\textbf{83.69\%}} (0.06) & \uline{\textbf{76.43\%}} (0.09) \\
\bottomrule
\end{tabular}}
\caption{Results of offline experiments. Metrics denoted in \textbf{bold} represent the best results in the unsupervised distillation phase, while \uline{\textbf{underscored and bolded}} denote the best results in the supervised distillation phase. All distilled results are average taken over 5 random seeds with standard deviation in parenthesis.}
\label{tab:results}
\end{table*}

\section{Experiments and Results}
\label{sec:bibtex}

\subsection{Dataset}
The questions and relevant web-pages we use are collected from a commercial search engine, the objective is to select a paragraph which can answer the question from the web-pages. We set question-paragraph pairs as samples need to be classified. Hundred millions of unlabeled pairs are collected for unsupervised distillation, and we obtained millions of labeled pairs which are used for teacher's fine-tune through crowd-sourcing annotators. The statistic of dataset is summarized in Table \ref{dataset}.

\subsection{Experiment Details}
\textbf{Teacher Architecture} In order to obtain multiple models with different structure and ability, we use the series of pretrained models ERNIE-2.0 \citep{sun2020ernie} with different layer and fine-tune them on different samplings of the total labeled data. The specific structural parameters for each teacher model can be found in Table \ref{tab:results}. Each model has been trained using a sample of 90\% of the total data for training purposes. 

\noindent \textbf{Student Architecture } Considering the computing resources and time consuming, we use the 12-layer transformer structure for online deployment.

In the training procedure, we use the Adam optimizer \citep{DBLP:journals/corr/KingmaB14} with $\beta_1 = 0.9$ and $\beta_2 = 0.99$. For all teacher models, we set the learning rate as 2e-5, the batch size as 64, and the warm-up step as 1000. The maximum length of input text is set as 384 and cross-entropy is used as loss function. In the distillation stage, we set the warm-up step as 1000, the learning rate as 2e-5 and the batch size as 64. The maximum length of input text is set as 384 and MSE is used as loss function. The best checkpoint is picked according to the performance on dev-set.

\subsection{Evaluation Metrics}
The metrics we used for experimental evaluation are introduced as below.

Precision-Recall is a useful measure of success of prediction when the classes are very imbalanced. $P = T_p/(T_p + F_p), R = T_p/(T_p + F_n)$ ,where $T_p$, $F_p$ and $F_n$ represent for the number of true positives, false positives and false negatives.

Different threshold of a classifier leads to different Precision-Recall, follow the need of online system, we take recall value where precision equals to 90\% as evaluation metrics.

\textbf{q R@P=90\%} This metric only takes the paragraph with highest predicted score among all candidates under given question into consideration. A question is noted as $T_p$ if the score of selected answer is higher than threshold and the label is positive, while $F_p$ means the score of selected answer is higher than threshold but the label is negative. If the score of selected answer is lower than threshold but it does exist a positive answer for this question, we note it as $F_n$. This question granularity metric follows the behavior of web-based OpenQA system since system only displays the best answer was found, so it can best imitate model's performance in online system.

\textbf{qp R@P=90\%} This metric takes every qp-pair sample into consideration so it can reflect model's general ability to find answers.

We also conduct a comparison called Good or Same or Bad (GSB) evaluation between two systems by inviting professional annotators to estimate which system produced a greater answer for each given question \citep{zhao-etal-2011-automatically}. The gain of a new system can be formulated as:
\label{GSB}
\begin{align}
\Delta_{GSB} & = \frac{\#Good - \#Bad}{\#Good + \#Same + \#Bad}
\end{align}

where \#Good (or \#Bad) denotes the number of questions that the new (or base) system provides better answer and \#Same denotes the number of questions that 
answer are equal in quality.

\textbf{r(query\_change)} The query change ratio, defined as the proportion of sessions where users initiate a subsequent search following their initial query, serves as an online user behavior metric. This study reports only the difference in the query change ratio between the experimental and baseline methods, withholding absolute values.

Lower query change ratio reflects better performance as users are satisfy with the initial response, obviating the necessity for further queries.

\begin{figure}
\centering
\includegraphics[scale=0.180]{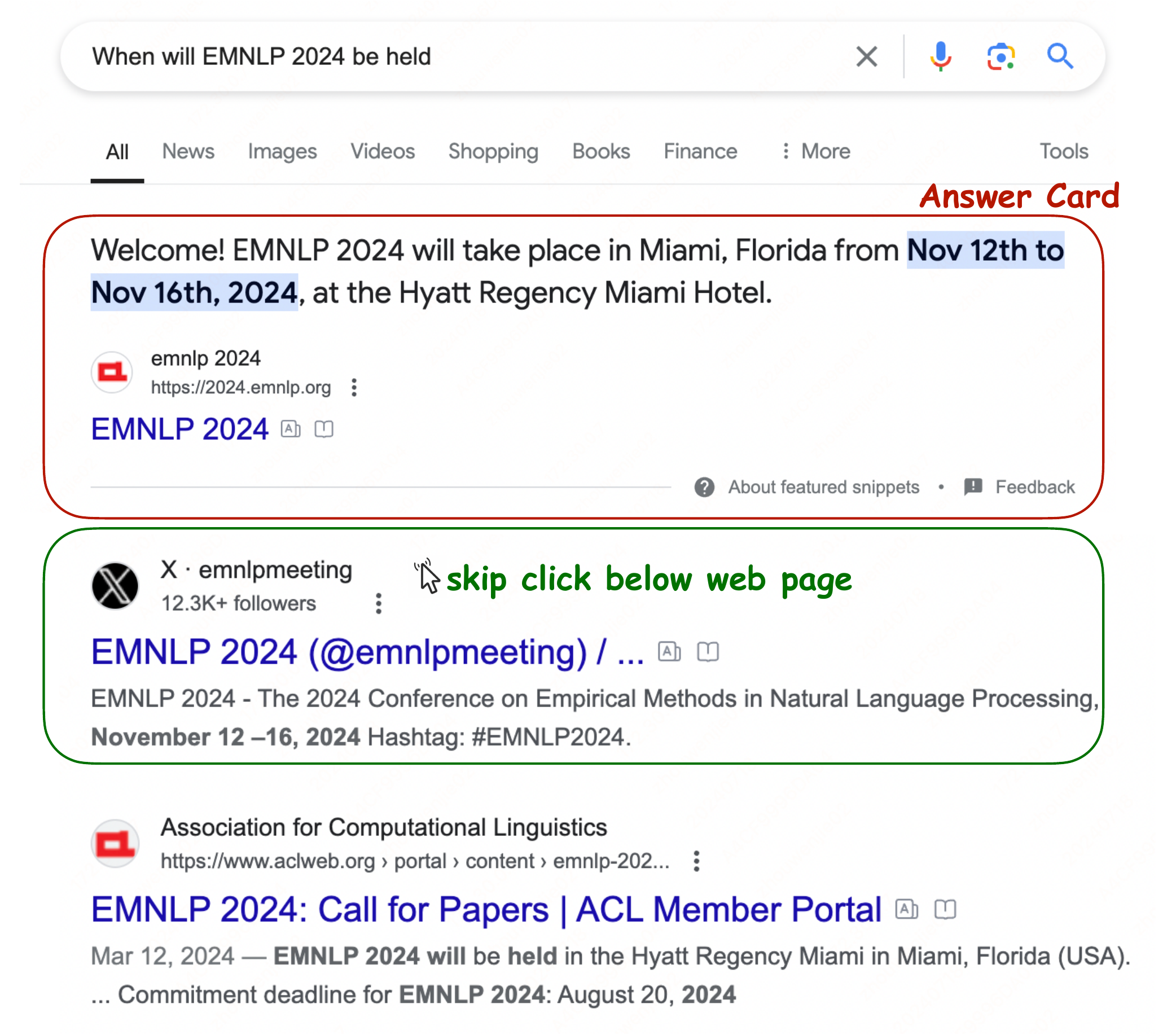}
\caption{\label{search_demo} The Answer Card is retrieve by the question answering system. Web pages below are not display in answer card format.}
\end{figure}

\textbf{r(skip\_click)} The skip click ratio, quantified as the proportion of instances where users click on web pages below the answer card (figure \ref{search_demo}), indicates potential dissatisfaction with the answer provided. Due to confidentiality constraints, we report only the differential in skip click ratios between the experimental and baseline methods. 

\subsection{Main Results}
The main results of distillation methods comparison are shown in Table \ref{tab:results}, we also display the results of teachers and ensemble methods. The methods used in the offline comparison experiments include:

\textbf{Mean Ensemble} We simply average the output of all teachers as the final predict score.

\textbf{Logistic Regression Weighted Ensemble} We trained a logistic regression model based on a dev-set to determine the weighting coefficients, and use these to obtain the weighted-sum of scores.

\textbf{MED}(Mean Ensemble Distillation) The predict score produced by Mean Ensemble Teachers is used as the optimizing object of student.

\textbf{LRED}(LR Ensemble Distillation) This variant uses Logistic Regression Weighted Ensemble Teachers for distillation instead of Mean Ensemble.

\textbf{CA-MKD} This an algorithm proposed by \citet{DBLP:conf/icassp/ZhangCW22} which adaptive assigns sample-wise reliability for each teacher prediction with the help of ground-truth labels, with those teacher predictions close to one-hot labels assigned large weights.

It is noted that the 125M distilled model outperforms the 10B teacher model. This could be attributed to the limited size of the training set, which comprises only 19K distinct questions and 2.5M labeled question-pair examples. Such a dataset may not be sufficiently large to leverage the full potential of the larger model. Additionally, an increase in the performance of the teacher models was noted throughout a finetuning epoch, suggesting that these models are underfitted. 

\subsection{Online Experiment}
To investigate the effectiveness of our proposed method in the real production environment, we deploy the proposed model in a commercial search engine, and conduct online experiments for comparison of MED and GOVERN.

\begin{table}[ht]
    \centering
    \scalebox{1.0}{
    \begin{tabular}{l|c|c}
    \hline
    \hline
     & Random & Tail \\
    \hline
        $\Delta_{GSB}$ & +4.5\% & +7.75\% \\
    \hline
        $G: S: B$ & 27: 364: 9 & 39: 353: 8 \\
    \hline
        $\Delta_{query\_change}$ & -0.68\% & -1.03\% \\
    \hline
        $\Delta_{skip\_click}$ & -3.46\% & -4.76\%\\
    \hline
    \hline
    \end{tabular}
    }
    \caption{Results of online experiments.}
    \label{MARCO_Results}
\end{table}

In contrast to random questions, tail questions are defined as those with a search frequency of less than 10 times per week. Given that heterogeneous search questions adhere to long-tail distributions, these tail questions constitute a significant portion of the questions processed by the search engine. It is evident that the proposed method consistently enhances the performance of the online QA system.

\subsection{Ablation Study}
Due to computational resource limitations, our ablation study utilized a 12-layer transformer as the teacher model and a 4-layer transformer as the student model. We divided the training data into ten folds, training each of the ten distinct teacher models on nine folds. The distillation process involved fifty million unlabeled samples, with the training epoch set to one.

The metric we report in this section is \textbf{qp prAUC}. This metric computes the area under the precision-recall curve where precision-recall is computed based on every qp-pair. It gives an overall measurement of classification ability.

\begin{figure}
\centering
\includegraphics[scale=0.180]{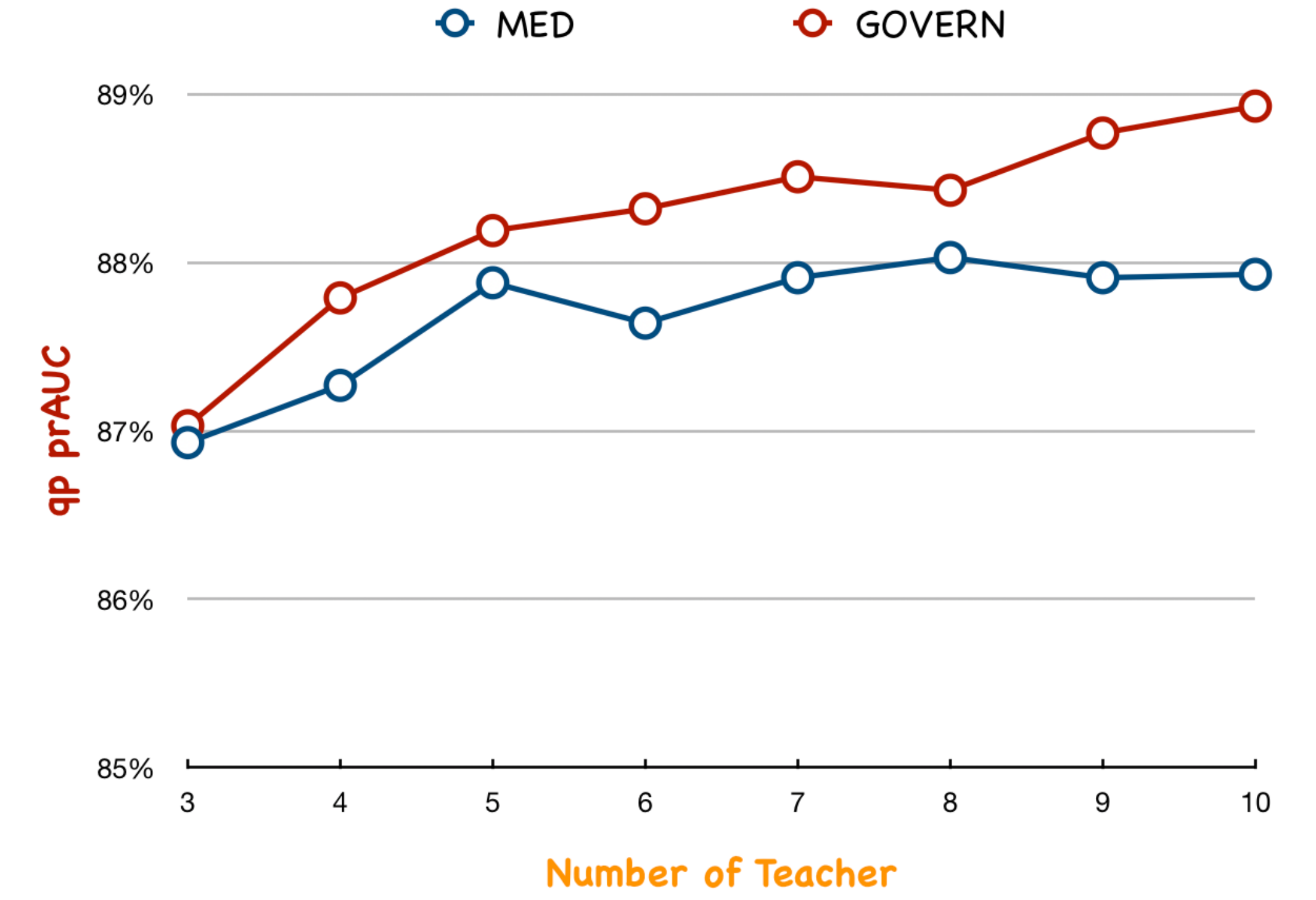}
\caption{\label{number} The effect of the number of teachers.}
\end{figure}

\textbf{Number of Teachers}
The impact of varying the number of teachers is illustrated in Figure \ref{number}. Experimental results indicate that the GOVERN algorithm consistently improves as the number of teachers increases. In contrast, mean-ensemble methods reach a performance plateau relatively quickly.

\textbf{Effect of Single Teacher}
We further investigate the impact of varying the performance of a single teacher, with results presented in Table \ref{single_Results}. The findings suggest that the GOVERN algorithm has the capacity to effectively select high-performing teachers, while simultaneously disregarding the noise generated by less effective ones.

\begin{table}[ht]
    \centering
    \scalebox{0.8}{
    \begin{tabular}{l|c}
    \hline
    \hline
     & qp prAUC \\
    \hline
        GOVERN with 5-teachers & 88.19\% \\
    \hline
        replace one teacher with 10B model & 89.03\% \\
        replace one teacher with 4-layer model & 88.11\% \\
    \hline
    \hline
    \end{tabular}
    }
    \caption{Effect of Single Teacher. }
    \label{single_Results}
\end{table}


\section{Related Work}

Following the seminal work of \citet{hinton2015distilling}, several studies have sought to develop advanced ensemble algorithms for distillation. We categorize these works into two groups based on their dependency on ground-truth labels.

\noindent \textbf{Unsupervised Ensemble Distillation} 
There are a few works focused on the ensemble method on unsupervised data (\citealp{DBLP:journals/corr/abs-1910-03581,sui-etal-2020-feded}), these works simply use the average output of multiple teachers as the distillation signal. Recently, \citet{DBLP:journals/corr/abs-2204-00548} and \citet{DBLP:conf/nips/IliopoulosKBMTV22} made efforts on distillation with unlabeled examples, but these studies primarily concentrate on dynamically assigning weight to unlabeled data. These approaches do not address the issue of teachers specializing in varying sample distributions.

\noindent \textbf{Supervised Ensemble Distillation} The idea of dynamic knowledge distillation with the help of ground-truth label was first explored by \citet{Du2020AgreeTD} and \citet{li-etal-2021-dynamic}. \citet{DBLP:conf/aaai/YuanSPLGFJ21} proposed a novel RL-based approach, which dynamically assigns weights to teacher models at instance level. \citet{cai-etal-2022-pile} proposed algorithm ensembles multi-teacher logits supervised by human-annotated labels in an iterative way. \citet{DBLP:conf/icassp/ZhangCW22} introduced conﬁdence-aware mechanism on both predictions and intermediate features for multi-teacher knowledge distillation.

\section{Conclusion}
In this paper, we present a novel algorithm, GOVERN, which dynamically selects teachers based on their gradient descent orientation. It does not require ground-truth labels, making it suitable for unsupervised distillation stages. Additionally, it can be integrated with existing supervised ensemble methods. The effectiveness of our method is affirmed through extensive experimentation.

\section*{Limitations}
The GOVERN algorithm does not currently account for the varying performance levels of teachers. This could be a shortcoming as it may be beneficial to assign a higher weight to more competent teachers, even if they share the same gradient descent orientation as other selected teachers.

As mentioned in section 3.1, existing dynamic methods are typically used to assign significance to samples, allowing GOVERN to integrate with them. We leave such integration as future work.

Theoretically, GOVERN is a general method that can be applied to other classification tasks. We conducted experiments specifically on the QA task because our team is responsible for the question-answering function in a search engine. We encourage readers to explore its application in different use cases.
\bibliographystyle{acl_natbib}
\bibliography{anthology,custom}


\appendix

\section{Appendix}
\label{sec:appendix}

In this section, we mathematically prove that the sample-wise dynamic weighting ensemble algorithm GOVERN is better than mean-ensemble. We only make the proof on positive samples, as for the negative samples, the proof process is the same due to the symmetry.

\subsection{Discrete Situation}

First, we consider the discrete case where each $teacher_i$ can be viewed as a classifier. For a binary classification model with precision of p, the probability of correct classification after each sampling follows a Bernoulli distribution. Thus, the expected classification precision of a single teacher is p, and the variance is p(1-p).

To simplify computation, we assume the performance of the N teachers is consistent, i.e., $p=p_1=...=p_N$, where $p_i$ is the precision of $T_i$. The mean ensemble of N teachers is formulated as:
\begin{align}
    X_{ME} = \frac{\Sigma_i X_i}{N}
\end{align}
given that $X_i$ which follows Bernoulli distribution are independent and identically distribute, we obtain the conclusion that $E(X_{ME}) = p$, $D(X_{ME}) = p(1-p)/N$.

Due to the fact $E(X_{ME}) = E(X_{M_i})$ and $D(X_{ME}) < D(X_{M_i})$, we conclude the following lemma: 

\textbf{Lemma 1.} \textit{Compared to the prediction from single model, although the mean ensemble result demonstrates better robustness, it keeping the expected precision the same.}

Next, we consider the case where N teachers form a vote-ensemble classifier based on the principle of maximum voting. Then the expectation of the classifier is as follows:
\label{unsupervised loss}
\begin{align}
p_0 = \sum_{m=\frac{N+1}{2}}^{N}C_{N}^{m}p^m(1-p)^{N-m}
\end{align}

Utilizing mathematical induction, it is trivial to prove when $p > 1/2$, $p_0 > p$. This is called Condorcet's jury theorem and details of proof can be found in \citep{DBLP:journals/mss/Sancho22}. Now we can state the following lemma:

\textbf{Lemma 2.} \textit{In discrete situation, vote-ensemble shows higher expected precision compared with mean-ensemble.}

\subsection{Consecutive Situation}

\begin{figure}
\centering
\includegraphics[scale=0.110]{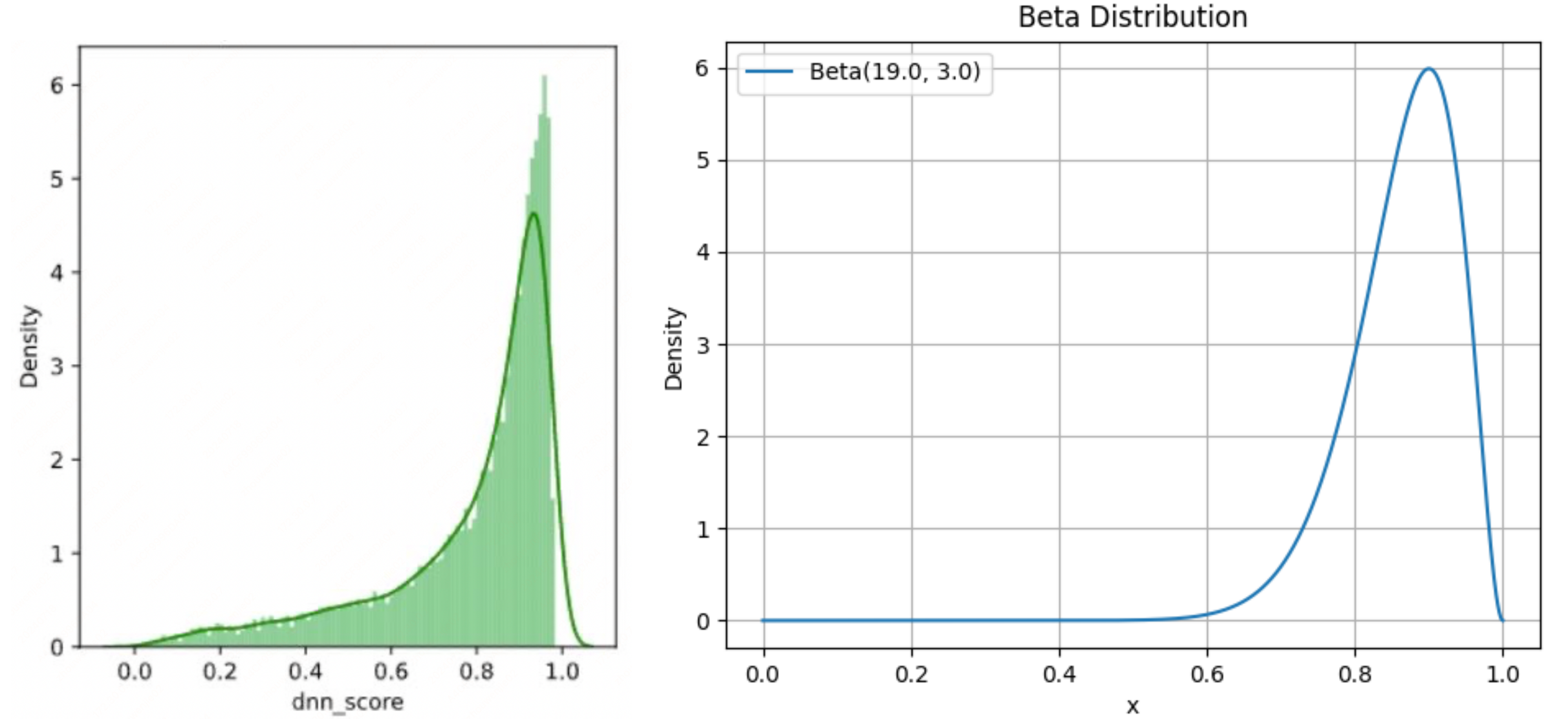}
\caption{\label{disribution} Left part shows the distribution of our model's output on test set, and right part shows the distribution of $Beta(19.0, 3.0)$. We can see that the model's output keep similar distribution with Beta function.}
\end{figure}

It is noted that in the setting of distillation, we take model as scorer rather than a simple classifier, and the output of the scorer is a float in $[0, 1]$. The distribution of the output is subject to Beta distribution, which is the conjugate distribution of Bernoulli distribution. This assumption can also be empirically verified as Figure \ref{disribution} shows. 

To simplify computation, we assume all teachers is subject to the same distribution, i.e., $X_i \sim B(b_1, b_2), \forall i \in \{1, ..., N\}$. Then we have:
\begin{align}
    E(X_{ME}) & = \frac{\sum E(X_i)}{n} = \frac{b_1}{b_1 + b_2} \\
    D(X_{ME}) & = \frac{\sum D(X_i)}{n^2} \\
    & = \frac{(b_1 * b_2)}{n * (b_1 + b_2)^2 * (b_1 + b_2 + 1)}
\end{align}
So \textit{Lemma 1} still holds in consecutive situation.

Next, we consider the case where N teachers calculate the ensemble scores by utilizing GOVERN method. We conduct numerical simulation using Monte-Carlo sampling to verify the superiority of GOVERN.

\begin{figure}
\centering
\includegraphics[scale=0.200]{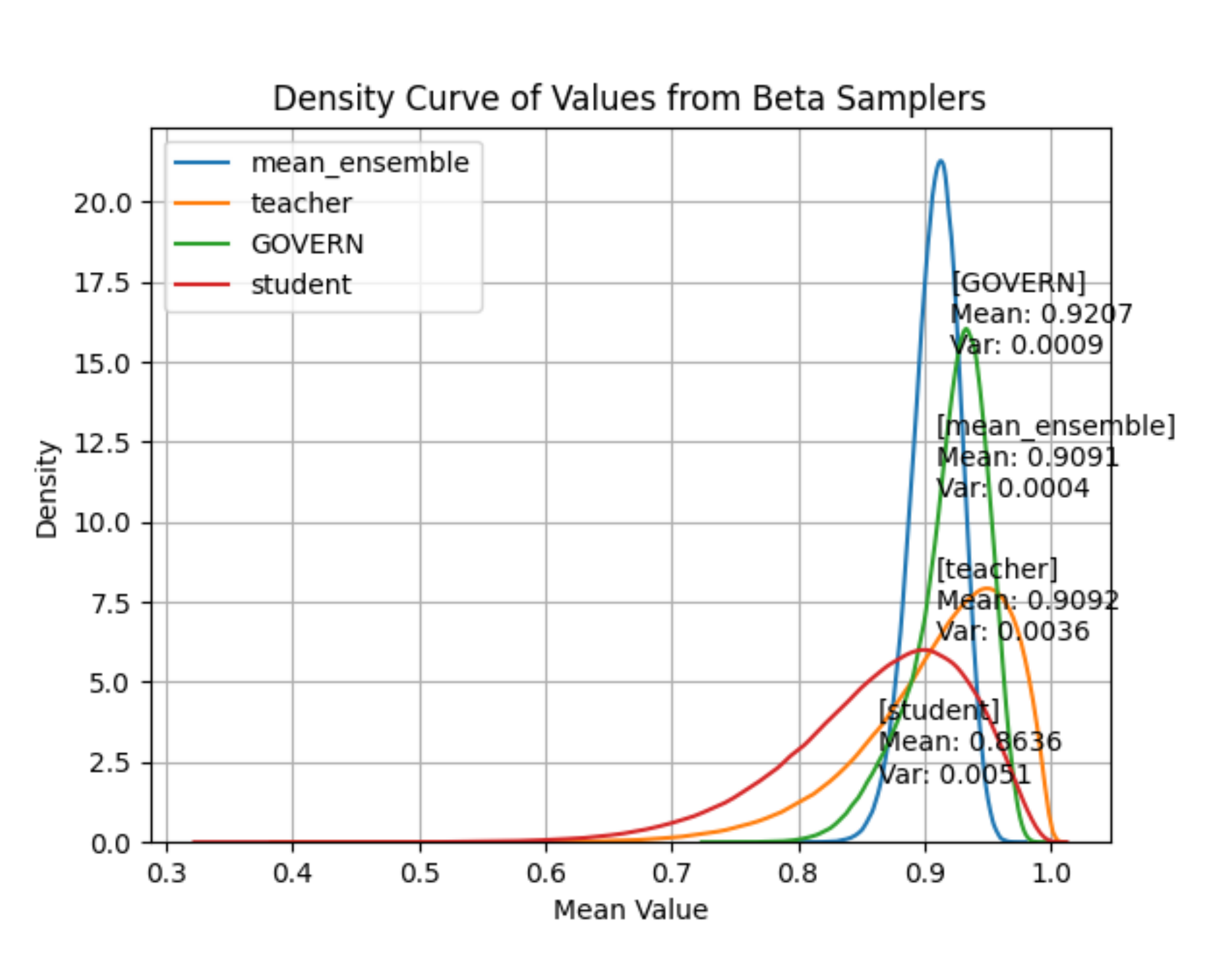}
\caption{\label{sampler}}
\end{figure}

We set 10 teachers with same distribution as $X_i \sim B(20.0, 2.0), \forall i \in \{1, ..., N\}$, and set student as $X_0 \sim B(19.0, 3.0)$. The number of simulation is set to 1M. 

The simulation result is shown in figure \ref{sampler}. We can see that the expectation of mean-ensemble is same with single teacher's output, while the variance is lower. This result is consist with \textit{Lemma 1}. Under the setting of GOVERN, it shows higher expectation compare with mean-ensemble, and keeps comparable variance. This verifies that GOVERN can obtain a better score with high expectation for distillation, and keep comparable robustness like mean-ensemble.


\end{document}